\pdfoutput=1

\documentclass[11pt]{article}

\usepackage{acl}

\usepackage{times}
\usepackage{latexsym}

\usepackage[T1]{fontenc}

\usepackage[utf8]{inputenc}

\usepackage{microtype}
\usepackage{amsmath}
\usepackage{graphicx}
\usepackage{multirow}
\usepackage{soul}

\usepackage{pifont}

\usepackage{times}
\usepackage{latexsym}
\usepackage{graphicx}
\usepackage{booktabs}
\usepackage{subcaption}
\usepackage[ruled,vlined]{algorithm2e}
\usepackage{multirow}
\usepackage{tabularx}
\usepackage{amsmath}
\usepackage{bbold}
\usepackage{mathtools}
\usepackage{xcolor}
\usepackage{booktabs}
\usepackage{tabularx,ragged2e}
\usepackage{enumitem}

\title{Can NLP Models Correctly Reason Over Contexts that Break the Common Assumptions?}

\author{First Author \\
  Affiliation / Address line 1 \\
  Affiliation / Address line 2 \\
  Affiliation / Address line 3 \\
  \texttt{email@domain} \\\And
  Second Author \\
  Affiliation / Address line 1 \\
  Affiliation / Address line 2 \\
  Affiliation / Address line 3 \\
  \texttt{email@domain} \\}

\author{Neeraj Varshney,~~ 
  Mihir Parmar,~~ 
  Nisarg Patel,~~ \\
  \textbf{Divij Handa},~~ 
  \textbf{Sayantan Sarkar}, ~~ 
  \textbf{Man Luo},~~   
  \textbf{Chitta Baral}
  \\
  Arizona State University \\
  }
\begin{document}
\maketitle
\begin{abstract}

Pre-training on large corpora of text enables the language models to acquire a vast amount of factual and commonsense knowledge which allows them to achieve remarkable performance on a variety of language understanding tasks.
They typically acquire this knowledge by learning from the pre-training text and capturing certain patterns from it.
However, real-world settings often present scenarios that do not abide by these patterns i.e. scenarios that break the common assumptions.
Can state-of-the-art NLP models correctly reason over the contexts of such scenarios?

Addressing the above question, in this paper, we investigate the ability of models to correctly reason over contexts that break the common assumptions.
To this end, we first systematically create evaluation data in which each data instance consists of (a) a common assumption, (b) a context that follows the assumption, (c) a context that breaks the assumption, and (d) questions based on the contexts.
Then, through evaluations on multiple models including GPT-3 and Flan T5, we show that while doing fairly well on contexts that follow the common assumptions, the models struggle to correctly reason over contexts that break those assumptions.
Specifically, the performance gap is as high as $20\%$ absolute points.
Furthermore, we thoroughly analyze these results revealing several interesting findings.
We believe our work and findings will encourage and facilitate further research in developing more robust models that can also reliably reason over contexts that break the common assumptions
\footnote{Data is available at \url{https://github.com/nrjvarshney/break_the_common_assumptions}}.





\end{abstract}

\begin{figure}[t!]
    \centering
    \includegraphics[width=7cm]{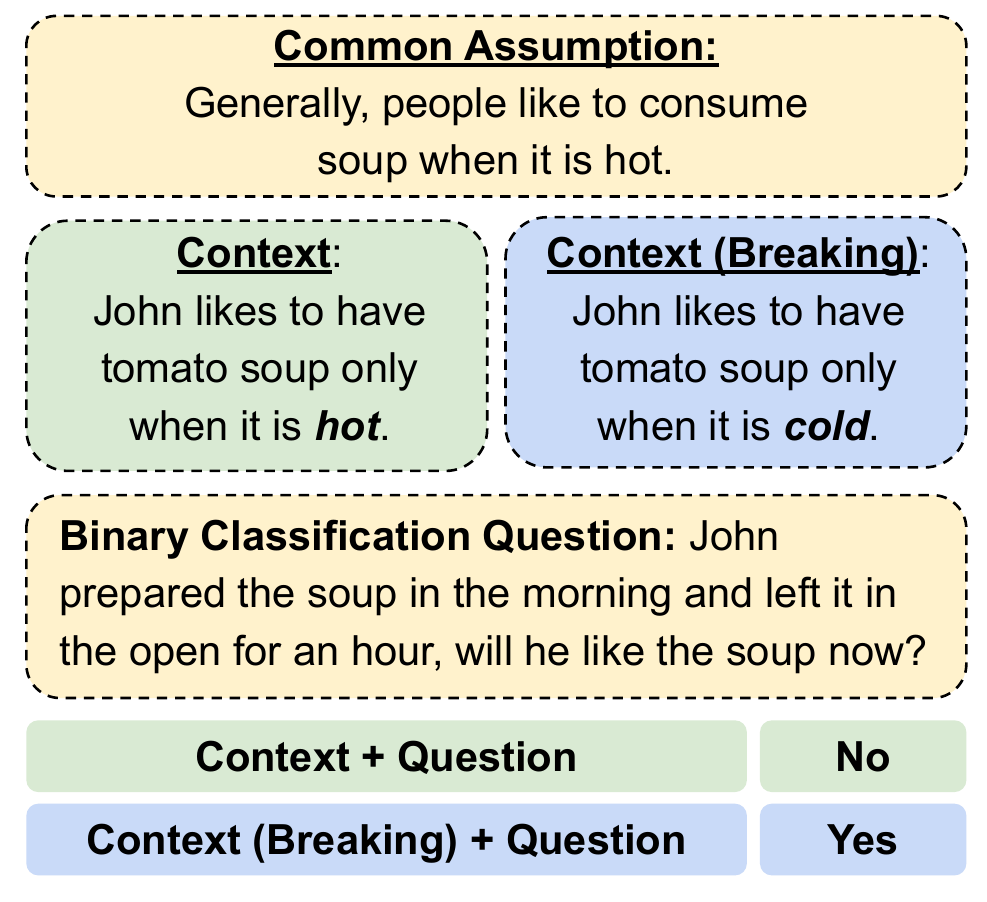}
    \caption{Illustrative examples of binary classification questions created in our study.
    \textbf{Context} follows the abovementioned common assumption while \textbf{Context (Breaking)} breaks it. 
    We show that state-of-the-art NLP models while performing well in reasoning over contexts that follow the common assumptions struggle to reason over contexts that break them.
    } 
    \label{fig:teaser_figure}
\end{figure}

\section{Introduction}



Pre-training on large corpora of text enables the natural language processing (NLP) models to acquire a vast amount of factual and commonsense knowledge \cite{liu-etal-2019-linguistic,petroni-etal-2019-language,yogatama2019learning,davison-etal-2019-commonsense}. 
Due to this knowledge, they are able to achieve remarkable performance on a variety of language understanding tasks.
They typically acquire this knowledge by learning from the pre-training text and capturing certain patterns from it.
However, in real-world settings, we often encounter scenarios that do not abide by these patterns i.e. scenarios that break the common assumptions.
Consider a context, `John likes to have tomato soup only when it is cold', this breaks the common assumption that `people prefer to consume soup when it is hot'.
Answering questions based on such contexts requires a model to truly understand the context and override its knowledge 
that it may have acquired (due to the predominant presence of certain patterns in the raw text) during pre-training.
How well can state-of-the-art NLP models perform in such scenarios?

Recently, many datasets have been created that test different language understanding skills such as pronoun resolution \cite{sakaguchi2021winogrande,levesque_winograd_2012}, commonsense reasoning \cite{talmor-etal-2019-commonsenseqa}, numerical reasoning \cite{dua-etal-2019-drop,patel-etal-2021-nlp,mishra-etal-2022-numglue}, qualitative reasoning \cite{tafjord-etal-2019-quartz, tafjord2019quarel}, temporal reasoning \cite{zhou-etal-2019-going}, and feasibility reasoning \cite{gupta2022john}. 
Furthermore, numerous adversarial datasets \cite{mccoy-etal-2019-right,bartolo-etal-2020-beat, naik-etal-2018-stress} have also been developed that test the robustness of models.
\citet{longpre-etal-2021-entity} study entity-based conflicts in the parametric and contextual knowledge.
\citet{agarwal2020entity} investigate entity-based swapping to test the robustness of models.
Prior work has also studied creating counterfactuals using various techniques such as token substitutions and adversarial attacks \cite{ribeiro-etal-2020-beyond,michel-etal-2019-evaluation,kaushik2020learning}.
However, evaluating models on the ability to reason over contexts that break the common assumptions (this is different from entity-based conflicts) has remained underexplored, and existing datasets do not contain a sufficient number of such examples.

In this work, we address the above limitations and comprehensively study the models' ability to reason over contexts that break the common assumptions.
To this end, we first systematically create questions (binary classification) in which the contexts break the common assumptions and the questions test the ability to reason over those contexts. 
Furthermore, for each such context, we also create a corresponding context that `follows' the common assumption.
Specifically, instances in our evaluation data consist of the following:
(a) a common assumption,
(b) a context that follows the assumption,
(c) a context that breaks the assumption, and
(d) questions based on the contexts.
Figure \ref{fig:teaser_figure} illustrates examples of our dataset.
For binary classification questions, the task is to answer a given question as either `Yes' or `No'.

We conduct comprehensive experiments with several NLP models such as Flan T5 \cite{chung2022scaling}, GPT-3 \cite{NEURIPS2020_1457c0d6}, and UnifiedQA \cite{khashabi-etal-2020-unifiedqa}.
First, we evaluate models on the scenario where the contexts follow the common assumptions; we show that the models perform fairly well in this setting.
\textbf{However, on evaluating them for the scenario where the contexts break the common assumptions, we find that the models falter and achieve considerably lower performance.}
Specifically, on the binary classification questions, Flan T5-xxl achieves an accuracy of just $70.67\%$ in the latter scenario ($\sim20$ absolute points lower than its performance on the former scenario).
\textbf{Furthermore, we show that this performance is considerably and consistently lower than the human performance baseline.}

We further conduct a thorough analysis which reveals several interesting findings such as (a) \textbf{models show poor consistency }i.e. they are often not able to correctly answer both (context-question) and (context (Breaking)-question) pairs correctly and (b) \textbf{explicitly providing the common assumption along with the context improves the performance when the context aligns withe the assumption but degrades when it breaks the assumption}.
Overall, we believe our work and findings will encourage and facilitate further research in developing more robust models that can also reliably reason over contexts that break the common assumptions. 


\begin{table*}[t]
    \small
    \centering
    \resizebox{\linewidth}{!}{
    \begin{tabular}{@{}p{0.1\linewidth}>{\RaggedRight}p{0.23\linewidth}>{\RaggedRight}p{0.23\linewidth}>{\RaggedRight}p{0.23\linewidth}}
    \toprule
        \textbf{Category} & 
        \textbf{Common Assumption} &
        \textbf{Context (Breaking)} &
        \textbf{Question (Binary Class.)}        
        \\
    \toprule
           \textbf{Preferences}	& 
          Generally, people prefer homes that are spacious and have adequate storage space.	& 
          John prefers small homes so that he can manage it properly.  &
          John's parents are looking for a new bungalow for him, will he like it? \textbf{No}\\ \\

          \textbf{Behaviors}	& 
          Generally, people feel good when they meet an old friend.	& 
          Kevin had a traumatic childhood because of which he feels uncomfortable meeting people from his growing up years.  &
          Kevin is invited for his school reunion celebration, will he enjoy the celebration and meeting his school friends? \textbf{No} \\ \\

          \textbf{Objects}	& 
          Generally, a branded watch is more expensive than a regular watch
          & 
          A premium brand which is known for expensive luxury watches has launched a new collection of watches which are available at low prices in this month. &
          Jimmy is looking to buy a watch but has a low budget, should he go for this premium brand? \textbf{Yes} \\ \\

          \textbf{Events}	& 
          Generally, sporting events have an audience & 
          This year soccer final is being held without an audience &
          The final soccer match will be played between the two most popular teams. Will the stadium be full of supporters of both teams? \textbf{No} \\  \\
          
          \textbf{Others}	& 
          Animals usually do not enter the gym. & 
          A gym in NY has a high membership fee and thus has no restrictions on the working hours and entry. &
          Will the gym allow Jim to take his dog with him? \textbf{Yes}\\

    \bottomrule

    \end{tabular}
    }
    \caption{Illustrative examples of binary classification questions corresponding to different categories of common assumptions. Correct answers are highlighted in \textbf{bold}.}
    \label{tab:examples}
\end{table*}

\section{Evaluation Data}
In order to comprehensively study a system's ability to reason over contexts that break the common assumptions, we first systematically create evaluation instances.
In this section, we describe the data creation process and provide supporting details.

\subsection{Data Creation}
\begin{table*}[t]
    \centering
    \small
    \begin{tabular}
    {@{}p{4.5cm}p{10.5cm}}
        \toprule
        \textbf{Context} & \textbf{Question}\\
        \midrule
        
        
        
        
        

        Matt always enjoys watching one-sided sports game & 
        Q1: There are two matches tonight. One is high-intensity close match. Other is a boring one-sided game. Will Matt watch the close match? \textbf{No}\\ \\
        
        & Q2: There are two matches tonight. One is high-intensity close match. Other is a boring one-sided game. Will Matt watch the one-sided match? \textbf{Yes}\\
        
        \midrule
        
        Matt doesn't enjoy watching interesting sports game but likes one-sided games & 
        Q1: There are two matches tonight. One is high-intensity close match. Other is a boring one-sided game. Will Matt watch the close match? \textbf{No}\\
        
        & Q2: There are two matches tonight. One is a high-intensity close match and the ather is a boring one-sided game. Will Matt prefer to watch the one-sided match? \textbf{Yes}\\

    \bottomrule
    \end{tabular}
    \caption{
    Illustrative examples of variations of a (context, question) pair in our dataset. This is used to comprehensively evaluate a system's ability to correctly and consistently answer questions.
    }
    \label{tab:variants_examples}
\end{table*}

For creating data instances, we first compile a set of common assumptions across various categories, namely assumptions about preferences, behaviors, objects, and events.
Table \ref{tab:examples} demonstrates examples of common assumptions for each category.
Then, we write a context that follows the common assumption and a corresponding context that breaks that assumption. 
Finally, we create binary classification questions from these contexts.
Furthermore, we also create several variants of a (context, question) pair to comprehensively evaluate a system's ability to correctly and consistently answer questions.
Table \ref{tab:variants_examples} shows examples of such variants. 
\textbf{We note that in this work, our focus is on common assumptions and not on entity-based factual knowledge. }

Six computer science graduate students contributed to the development of this dataset. 
The data instances were cross-verified and instances on which the inter-annotator agreement was low were rejected.
We also conduct validation of the compiled common assumptions; specifically, for each sentence, we asked human annotators to answer `Yes' if they think that it is a common assumption otherwise answer `No'.
For nearly all the compiled common assumptions, the majority answer is `Yes' which posits that they are indeed common assumptions.
We provide further details about this step in section \ref{sec_data_validation}.

\paragraph{Categories of common assumptions:}
We create common assumptions for the following categories:

\textbf{Assumptions about Preferences}: In this category, we include assumptions where a preference (typically of humans) is involved; for e.g. ``\textit{Generally, people prefer to eat fruits when they are fully ripened}'', ``\textit{Generally, busy people prefer to have an assistant who can help them with their tasks}'', and ``\textit{People usually like to go outside when the weather is pleasant}''.

\textbf{Assumptions about Behaviors}: Here, we include assumptions about people's behaviors such as `Generally, people feel good when they meet an old friend', `Generally, people like to get free coupons', and `People usually go to work in the morning'.

\textbf{Assumptions about Objects}: This category incorporates assumptions about objects/things such as `Generally, hotels are more expensive than a dormitory', `Generally, bigger vehicles have more seating capacity', and `Generally, schools have science laboratories'.

\textbf{Assumptions about Events}: In this category, we include assumptions about events such as `Generally, football games have an audience' and `Generally, there are food stalls in a carnival celebration'.

We also include an \textbf{Others} category to incorporate common assumptions that do not fit into the above four categories.




\begin{table}[t]
    \centering
    \begin{tabular}
    {@{}p{4.5cm}p{2.5cm}}
        \toprule
        \textbf{Category (\# Assumptions)} & \textbf{\# Questions} \\
        \midrule

        Preferences (33) & 131 \\
        Behaviors (64) & 240 \\
        Objects (17) & 73 \\
        Events (26) & 95 \\
        Others (13) & 44 \\

    \bottomrule
    \end{tabular}
    \caption{Number of binary classification questions for each category in our evaluation data.
    }
    \label{tab:common_assumptions}
\end{table}



\subsection{Data Statistics}
For binary classification questions, the task is to answer a given question as either `Yes' or `No'.
To further measure the consistency of a system's predictions, we evaluate its predictions on context pairs where one context follows the common assumption and a corresponding context that breaks it.
We also conduct evaluations on different variations of a (context, question) pair as shown in Table \ref{tab:variants_examples}.
Table \ref{tab:common_assumptions} shows the number of binary classification questions in our dataset across each category.




\begin{table*}[t]
    \centering
    \resizebox{0.97\linewidth}{!}{
    \begin{tabular}
    {@{}c|cc|cc|cc|cc|cc|cc}
        \toprule

        \textbf{Model} &
  \multicolumn{2}{c}{\textbf{Preferences}} &
  \multicolumn{2}{c}{\textbf{Behaviors}} &
  \multicolumn{2}{c}{\textbf{Objects}} &
  \multicolumn{2}{c}{\textbf{Events}} &
  \multicolumn{2}{c}{\textbf{Others}} & 
  \multicolumn{2}{c}{\textbf{Average}}\\

  & Con & Con (B) & Con & Con (B) & Con & Con (B) & Con & Con (B) & Con & Con (B) & Con & Con (B) \\
  
        \midrule
        \textbf{Human} & 95.00 & 95.00 & 100.00 & 95.00 & 100.00 & 95.00  & 100 & 95.00 & 100.00 & 100.00 & 99.00 & 96.00 \\
        
        \midrule
        

        \textbf{Flan T5-xxl} & 88.17 &  68.7 &  90.21 &  69.79 &  89.04 &  71.91 &  89.47 &  77.89 &  86.36 &  63.64 &  89.19 &  70.67 \\
        

        \textbf{Flan T5-xl} & 85.12 &  72.52 &  90.21 &  66.25 &  86.98 &  69.17 &  88.95 &  70.0 &  81.82 &  65.91 &  87.83 &  68.61 \\


        \textbf{Flan T5-large} & 87.03 &  66.03 &  82.5 &  61.05 &  79.45 &  67.12 &  92.63 &  63.68 &  81.82 &  62.5 &  84.73 &  63.47  \\
        

        \textbf{Flan T5-base} & 66.03 &  56.49 &  71.66 &  58.96 &  63.69 &  60.95 &  80.53 &  58.42 &  67.05 &  55.69 &  70.5 &  58.32 \\
        

        \textbf{UnifiedQA} & 63.36 &  50.77 &  66.04 &  46.66 &  60.27 &  52.73 &  65.26 &  54.21 &  60.23 &  51.14 &  64.15 &  49.91 \\
        



        \textbf{GPT-3 davinci-003} & 90.84 & 56.49 & 88.75 & 54.17 & 78.08 & 57.53 & 88.42 & 54.74 & 84.09 & 72.73 & 87.48 & 56.6 \\
        

    \bottomrule
    \end{tabular}
    }
    \caption{
    Performance of different models on binary classification questions for each category of our evaluation data. 
    Column `Con' and column `Con (B)' correspond to the performance on context-question pairs where contexts follow the common assumptions and  where contexts break the assumptions respectively.
    }    
    \label{tab:binary_results}
\end{table*}

\subsection{Data Validation}
\label{sec_data_validation}
We note that it is important to validate the quality of the compiled common assumptions. 
To this end, for each sentence, we ask 3 human annotators to answer `Yes' if they think that the given sentence is a common assumption otherwise answer `No'.
Then, we use the majority voting aggregation strategy and find that for nearly all the compiled common assumptions, the majority answer is `Yes'. 
This validates the quality of the common assumptions compiled in this work.

In addition to the above validation step, we note that the questions were also cross-verified by the data creators (who are also the authors of this paper) and the instances where the inter-annotator agreement was low were rejected.


\section{Experiments}

\subsection{Experimental Setup}
\paragraph{Performance Metrics:} 
For binary classification questions, the task is to answer a given question as either `Yes' or `No'.
We calculate accuracy against the gold labels (Yes and No) for evaluation. 
To better evaluate a system’s capability, we measure its consistency in correctly answering both the scenarios corresponding to the context that follows the common assumption and the context that breaks it.

\paragraph{Models:}
We evaluate Flan T5 \cite{chung2022scaling}, GPT-3 (text-davinci-003) \cite{NEURIPS2020_1457c0d6}, and UnifiedQA \cite{khashabi-etal-2020-unifiedqa} models on our task. 

\paragraph{Human Performance Baseline: }
We randomly select 40 context-question pairs (20 for contexts that follow common assumptions and 20 for corresponding contexts that break those assumptions) for each category and ask a total of 3 human annotators to `answer the given question in Yes or No based on the context'.
We then use the majority voting aggregation method and calculate the human performance baseline.


\subsection{Results}

Table \ref{tab:binary_results} shows the performance of different models on binary classification questions. Column `Con' and column `Con (B)' correspond to the performance on context-question pairs where contexts follow the common assumptions and  where contexts break the assumptions respectively.

\paragraph{High Human Performance Baseline:}
The first row in Table \ref{tab:binary_results} shows the human performance baseline for each category of our evaluation data.
It demonstrates that humans typically achieve high performance across all the data categories. 
This shows that typically humans are able to reason well in both the scenarios i.e. where contexts follow the common assumptions and where contexts break those assumptions.
On average, the human performance is $99\%$ on `Con' and $96\%$ on `Con (B)'.

\paragraph{Con vs Con (B) Performance: }
On comparing the performance on questions for contexts that follow the common assumptions (`Con') and for contexts that break them (`Con (B)'), we find that the models consistently achieve lower performance on `Con (B)'.
This behavior is observed for all the models and for all categories of common assumptions.
For instance, Flan T5-xxl model on average achieves $89.19\%$ accuracy on `Con' and just $70.67\%$ on `Con (B)'.
The gap in performance is observed for all the categories of common assumptions.
The table also shows that with the increase in the size of the model, the performance on both `Con' and `Con (B)' improves. However, the gap in performance on them remains.
\textbf{This highlights that despite performing fairly well on reasoning over the contexts that follow the common assumptions, the models struggle to correctly reason over contexts that break those common assumptions.}

\paragraph{Human vs Model Performance on `Con (B)':}
Table \ref{tab:binary_results} shows that the performances of all models are considerably lower than the human performance baseline. 
Specifically, on `Con (B)' instances, the human performance on average is $\sim 26\%$ higher than the Flan T5-xxl model.
\textbf{Furthermore, human performance is just slightly impacted when the contexts break the common assumptions (i.e. `Con' column); however, the models' performance degrades significantly. }
This behavior is observed for all the categories.

\paragraph{Models Show Poor Consistency:}
\begin{table}[t]
    \centering
    \resizebox{0.97\linewidth}{!}{
    \begin{tabular}
    {@{}c|ccccc}
        \toprule
        \textbf{Model} & \textbf{Pref} & \textbf{Beh} & \textbf{Obj} & \textbf{Eve} & \textbf{Oth}\\
        
        \midrule
        
        \textbf{Flan T5-xxl} & 56.3 & 58.68 & 58.9 & 68.42 & 50.0\\
        
        \textbf{Flan T5-xl} & 54.81 & 54.96 & 56.16 & 55.79 & 45.45 \\
        
        \textbf{Flan T5-large} & 47.41 & 40.91 & 42.47 & 48.42 & 34.09 \\

        \textbf{Flan T5-base} & 24.44 & 32.23 &  26.03 & 35.79 & 20.45\\


        \textbf{UnifiedQA} & 28.89 & 19.42 & 21.92 & 26.32 & 15.91 \\


        \textbf{GPT-3} & 49.62 & 47.08 & 41.1 & 46.32 & 56.82 \\

    \bottomrule
    \end{tabular}
    }
    \caption{
    Consistency of different models on the binary classification questions.
    }
    \label{tab:binary_consistency}
\end{table}
Table \ref{tab:binary_consistency} shows the consistency (correctly answering a question based on both Context and Context (Breaking)) achieved by different models on the binary classification questions.
The results show that all the models achieve poor consistency i.e. they are often not able to correctly answer both (context-question) and (context (Breaking)-question) pairs correctly.
This is primarily due to the poor performance on (context (Breaking)-question) instances.

\begin{table*}[t]
    \centering
    \resizebox{0.97\linewidth}{!}{
    \begin{tabular}
    {@{}c|cc|cc|cc|cc|cc|cc}
        \toprule

        \textbf{Model} &
  \multicolumn{2}{c}{\textbf{Preferences}} &
  \multicolumn{2}{c}{\textbf{Behaviors}} &
  \multicolumn{2}{c}{\textbf{Objects}} &
  \multicolumn{2}{c}{\textbf{Events}} &
  \multicolumn{2}{c}{\textbf{Others}} & 
  \multicolumn{2}{c}{\textbf{Average}}\\

  & Con & Con (B) & Con & Con (B) & Con & Con (B) & Con & Con (B) & Con & Con (B) & Con & Con (B) \\
  
        \midrule

        \textbf{Flan T5-xxl} & 91.6 & 65.65 & 92.08 & 70.83 & 93.15 & 61.64 & 93.68 & 73.68 & 88.64 & 59.09 & 92.11 & 68.1 \\

        \textbf{Flan T5-xl} & 87.79 & 64.89 & 92.08 & 63.33 & 89.04 & 63.01 & 90.53 & 60.0 & 88.64 & 54.55 & 90.22 & 62.44 \\

        \textbf{Flan T5-large} & 84.73 & 66.41 & 87.08 & 60.42 & 80.82 & 46.58 & 86.32 & 58.95 & 84.09 & 56.82 & 85.42 & 59.52 \\

        \textbf{Flan T5-base} & 64.12 & 54.96 & 75.0 & 52.92 & 65.75 & 60.27 & 78.95 & 48.42 & 65.91 & 45.45 & 71.36 & 53.0 \\ 

        \textbf{UnifiedQA} &  66.41 & 38.93 & 68.33 & 44.58 & 57.53 & 50.68 & 65.26 & 47.37 & 54.55 & 47.73 & 65.01 & 44.77 \\

    \bottomrule
    \end{tabular}
    }
    \caption{
    Performance of different models on binary classification questions (when the common assumption is explicitly provided with the context) for each category of our evaluation data. 
    Column `Con' and column `Con (B)' correspond to the performance on context-question pairs where contexts follow the common assumptions and  where contexts break the assumptions respectively.
    }    
    \label{tab:binary_results_with_assumptions}
\end{table*}
\paragraph{Impact of Explicitly Providing the Common Assumption with the Context:}

Table \ref{tab:binary_results_with_assumptions} shows the impact of explicitly providing the common assumption along with the context.
Since the common assumption aligns with the `Con' contexts, it slightly improves the performance on `Con'; however, it hurts the performance on `Con (B)'.
This happens because the contexts in `Con (B)' break the provided common assumptions. Hence, it further distracts the model resulting in a drop in performance.


\paragraph{Failure Instances:}

\begin{table*}[t]
    \small
    \centering
    \resizebox{\linewidth}{!}{
    \begin{tabular}{@{}p{0.35\linewidth}>{\RaggedRight}p{0.45\linewidth}>{\RaggedRight}p{0.1\linewidth}}
    \toprule
        \textbf{Context (Breaking)} &
        \textbf{Question (Answer)} &
        \textbf{Prediction} \\
        
    \toprule

          Ronald never hires a person that is experienced to handle his business.
          & 
          Joan is an inexperienced candidate applying for the position. will he be considered for hiring? 
          (\textbf{Yes})
          & 
        No \\ \\

        John is content with his small apartment and wants to continue to stay here
        & His parents offered to help him buy a bigger home, will he decline the offer?
        (\textbf{Yes})
        & 
        No \\ \\


         John enjoys in small homes so that he can manage it properly 
        & 
        John's parents are looking for a new bungalow for him, will he like it?
        (\textbf{No}) 
        & 
        Yes \\ \\

        Steven's has an old car that is even slower than a bicycle
        & 
        Steven rides his bicycle and car for one hour, will he cover more distance with bicycle?
        (\textbf{Yes}) 
        & 
        No \\ \\ 

        Matt always enjoys watching boring sports game
        & 
        There are two matches tonight. One is high intensity close match. Other is a boring one-sided game. Will Matt watch the one-sided match?
        (\textbf{Yes})
        & 
        No \\


    \bottomrule

    \end{tabular}
    }
    \caption{Examples of errors in prediction made by Flan T5-xxl model on the binary classification questions.}
    \label{tab:errors}
\end{table*}

Table \ref{tab:errors} shows examples of instances where Flan T5-xxl model gave incorrect predictions. 
On analyzing the failure instances, we find that a large fraction of the mistakes are on the instances where the correct answer is `Yes' while the model gives `No' as its prediction.

\paragraph{Performance on instance variations:}
\begin{table}[t]
    \centering
    \begin{tabular}
    {@{}cc}
        \toprule
        \textbf{Model} & \textbf{Performance} \\
        \midrule
        Flan T5-xxl  & 33.99 \\
        Flan T5-xl & 31.7 \\
        Flan T5-large & 21.57 \\
        Flan T5-base & 15.36 \\

    \bottomrule
    \end{tabular}
    \caption{
    Performance of different models on different variations of (context (Breaking)-question) pairs.
    }
    \label{tab:perf_on_variants}
\end{table}

Table \ref{tab:perf_on_variants} shows the overall performance of different models on different variations of (context (Breaking) - question) pairs i.e. if a model predicts all the variants corresponding to a common assumption correctly then we give it a score of 1 otherwise we give it 0.
Flan T5-xxl achieves a performance of just $33.99\%$ on this metric highlighting \textbf{that the model is often not able to consistently answer ALL the variants correctly}.


\section{Conclusion}
In this paper, we investigated the ability of models to correctly reason over contexts that break the common assumptions. 
To this end, we first systematically developed evaluation data that consists of a common assumption, a context that follows that assumption, a context that breaks the assumption, and question based on the contexts.
Then, we evaluated multiple models and show that while performing fairly well on contexts that follow the common assumptions, the models struggle to correctly reason over contexts that break those assumptions.
Furthermore, we conducted a thorough analysis which resulted in several interesting findings.
In conclusion, we believe our work and findings will encourage and facilitate further research in developing more robust models that can also reliably reason over contexts that break the common assumptions. 


\section*{Ethical Considerations}

The names used in our data are selected from the most common English names. 
Though the contexts in our dataset break the common assumption, we ensure that all of them indeed describe a realistic scenario.
We do not collect any personal information from data creators in the development of the evaluation data for this work. 

\section*{Acknowledgement}
We thank the Research Computing (RC) at Arizona State University (ASU) for providing computing resources for experiments.

\bibliography{anthology,custom}
\bibliographystyle{acl_natbib}






\end{document}